\documentclass[lettersize,journal]{IEEEtran}
\usepackage{lineno,hyperref}
\usepackage{multirow}
\usepackage{booktabs}
\usepackage[cmex10]{amsmath}
\usepackage{bm}
\usepackage{times}
\usepackage{epsfig}
\usepackage{graphicx}
\usepackage{caption}
\usepackage{amsmath}
\usepackage{amssymb}
\usepackage{caption}    
\graphicspath{{media/}} 
\newcommand{\etal}{\textit{et al.}}

\author{
Arbish Akram$^{*}$, 
Nazar Khan$^{*}$, 
Arif Mahmood$^{\dagger}$ \\
\\
$^{*}$Department of Computer Science, University of the Punjab, Lahore, Pakistan \\
$^{\dagger}$Department of Computer Science, Information Technology University, Lahore, Pakistan
}

\date{}

\begin{document}
\bstctlcite{IEEEexample:BSTcontrol}
    \title{Improving Generative Adversarial Network Generalization for Facial Expression Synthesis}
    \author{
Arbish Akram$^{*}$, 
Nazar Khan$^{*}$, 
Arif Mahmood$^{\dagger}$ \\
$^{*}$Department of Computer Science, University of the Punjab, Lahore, Pakistan \\
$^{\dagger}$Department of Computer Science, Information Technology University, Lahore, Pakistan
}
\markboth{}{}

\maketitle
\begin{abstract}
Facial expression synthesis aims to generate realistic facial expressions while preserving identity. Existing conditional generative adversarial networks (GANs) achieve excellent image-to-image translation results, but their performance often degrades when test images differ from the training dataset. We present Regression GAN (RegGAN), a model that learns an intermediate representation to improve generalization beyond the training distribution. RegGAN consists of two components: a regression layer with local receptive fields that learns expression details by minimizing the reconstruction error through a ridge regression loss, and a refinement network trained adversarially to enhance the realism of generated images. 
We train RegGAN on the CFEE dataset and evaluate its generalization performance both on CFEE and challenging out-of-distribution images, including celebrity photos, portraits, statues, and avatar renderings. 
For evaluation, we employ four widely used metrics: Expression Classification Score (ECS) for expression quality, Face Similarity Score (FSS) for identity preservation, QualiCLIP for perceptual realism, and  Fr\'echet Inception Distance (FID) for assessing both expression quality and realism. RegGAN outperforms six state-of-the-art models in ECS, FID, and QualiCLIP, while ranking second in FSS. Human evaluations indicate that RegGAN surpasses the best competing model by 25\% in expression quality, 26\% in identity preservation, and 30\% in realism.
\end{abstract}


\section{Introduction}\label{intro}
Facial expression synthesis aims to transform a given expression into a target one. Significant progress has been achieved, particularly with Generative Adversarial Networks (GANs) \cite{mirza2014conditional, karras2017progressive} and, more recently, diffusion models \cite{ho2020denoising, brooks2023instructpix2pix, brack2024ledits++}. GANs enable efficient adversarial training for controllable image generation, while diffusion models rely on iterative denoising to achieve higher fidelity and diversity. Among GAN variants, StyleGAN \cite{karras2019style, karras2021alias} has been widely used for facial expression manipulation. The input image is first projected into the StyleGAN latent space, where semantic editing is performed. The modified latent code is then mapped back to the image space. 
Despite these advances, the key challenge in FES is to satisfy these three requirements simultaneously: i) expression correctness - the synthesized image should accurately reflect the target expression, ii) identity preservation - the subject's identity should remain unchanged, and iii) facial detail preservation - the fine details of the input image should be retained. However, existing approaches struggle to meet all these requirements at once.

\begin{figure}[t]
    \centering
    \includegraphics[width=\linewidth]{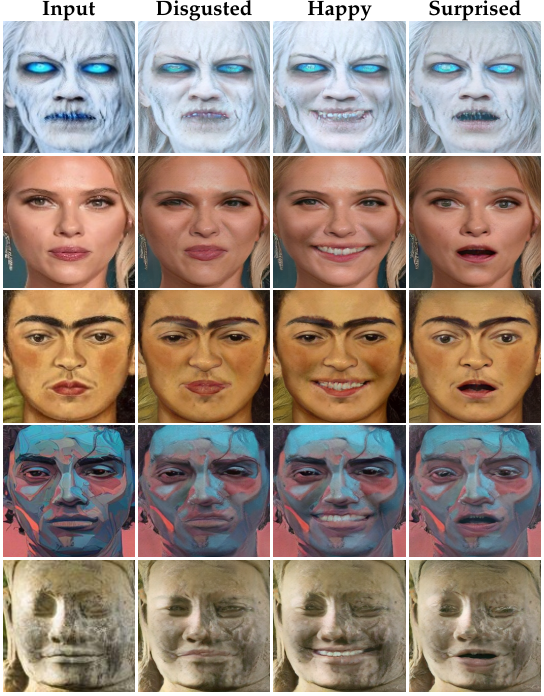}
    \caption{Given a neutral input image, the proposed RegGAN synthesizes photorealistic facial expressions on out-of-distribution testing images. Despite being trained only on real human faces from the CFEE dataset \cite{du2014compound}, the proposed method introduces realistic expressions, preserves identity and retains facial details of input images.}
    \label{fig:teaser_fig}
\end{figure}

In this work, we propose Regression GAN (RegGAN), a framework that combines regression and adversarial refinement for facial expression synthesis. A regression layer trained with least-squares objective generates coarse expression predictions, which are subsequently refined by a GAN to improve realism and preserve facial details.
 
The vanilla regression layer scales poorly with image size due to the rapidly increasing number of parameters. To address this, we employ a sparsified ridge regression layer that operates on local patches, significantly reducing the number of parameters while producing an intermediate representation of the desired expression. The refinement network is an encoder-decoder that takes this intermediate representation and passes it through three types of spatial attention blocks at multiple scales to enhance realism and preserve facial details. By using spatial attention instead of vanilla residual block, the network better captures expressions and subtle facial features. 
RegGAN is trained sequentially in two stages. First, the regression layer generates intermediate representations encoding the desired expressions. These generated images are then passed through the refinement network, which is trained with an adversarial loss to further enhance realism. This sequential training enables RegGAN to achieve superior performance in expression accuracy, identity preservation, and facial detail compared to existing methods. We evaluate RegGAN on out-of-distribution facial images, including portraits, avatars, statues, and animals, demonstrating its ability to induce human expressions in any face-like input.

\noindent The main contributions of this work are:
\begin{itemize}
    \item Augmented a localized variant of regression with adversarial learning. This tailored regression conditioning significantly improves the generalization ability of GANs, particularly on challenging image types such as portraits, avatars, fantasy artworks, illustrations, impasto paintings, and statues.
    \item Used a refinement network with multi-scale attention residual blocks to enhance both global and local facial features, improving the realism and fidelity of generated expressions.
    \item Adopted a sequential training strategy for the regression and refinement networks, enabling robust performance on out-of-distribution face-like inputs.
\end{itemize}

The paper is organized as follows. Section \ref{sec:related_work} provides a brief overview of generative adversarial networks and state-of-the-art facial expression synthesis models. Section \ref{sec:proposed_method} describes the details of our proposed method. Dataset and implementation details as well as qualitative and quantitative results and ablation studies, of the proposed method are provided in Section \ref{sec:exp_evaluation}. Section \ref{sec:discussion} provides insights into the strengths and weaknesses of the proposed method, while Section \ref{sec:failure_cases} discusses its failure cases. Section \ref{sec:conclusion} concludes the paper.

\section{Related Work}
\label{sec:related_work} 
Recently, with the advent of GANs, photo-realistic results have been achieved in image-to-image translation tasks \cite{isola-2016, zhu-2017,choi-2017,kim-2017,huang2017beyond,di2017gp, huang2021multi, rashid2023high, zhao2020fine}. Pix2pix \cite{isola-2016} introduced a conditional adversarial network with $\ell_1$ and adversarial losses to learn mapping between paired domains.
To address the problem of paired data alignment, CycleGAN \cite{zhu-2017} enabled unsupervised image-to-image translation via cycle-consistency. Both methods have been applied to facial expression synthesis \cite{khan_mr_ijcv_2020}, but their performance degrades on out-of-distribution images.

StarGAN \cite{choi-2017} enables multi-domains facial expression synthesis through a single generator-discriminator architecture. GANimation \cite{pumarola2018ganimation} generates facial expressions of varying intensity for realistic animation but requires action unit as input at inference.
Cascade-EFGAN \cite{wu2020cascade} incorporates local and global attention mechanisms, along with a progressive editing strategy, to better preserve facial details and reduce artifacts. 
Domain-adaptive translation has also been explored to generate expressions across diverse image styles such as sketches, oil paintings, and animal faces \cite{chen2020domain}. More recently, USGAN \cite{akram23usgan} and SARGAN \cite{akram23sargan} introduced an ultimate skip connection and spatial attention mechanism within residual blocks to better preserve facial details. Despite its benefits, the ultimate skip connection can sometimes be ineffective in inducing expressions.

All of these GAN-based methods produce good results when the testing images are similar to those on which they were trained. However, their performance may degrade when applied to test images taken from the wild. This is because each training dataset has its own bias, which is also learned by the GAN \cite{herranz2016scene,torralba2011unbiased}. Such dataset bias limits the generalization capacity of the GANs.

StyleGAN-based \cite{karras2019style, karras2021alias} latent manipulation methods have also been explored for facial editing \cite{wang2022high, hu2022style, abdal2020image2stylegan++}. Input images are first inverted into the latent space of StyleGAN, then latent codes are edited for facial manipulation. Authors in \cite{tov2021designing, hu2022style, alaluf2022hyperstyle} proposed inversion techniques to improve the reconstruction of input images with fine facial details. E4e \cite{tov2021designing} learns an encoder that balances reconstruction fidelity and editability. StyleTransformer \cite{hu2022style} leverages a transformer architecture to enhance latent inversion quality. HyperStyle \cite{alaluf2022hyperstyle} employs hypernetworks to predict image-specific adaptations to the StyleGAN generator, achieving both fidelity and editability. StyleRes \cite{pehlivan2023styleres} introduces residual-based feature transformations to better capture fine details during inversion. StyleFeatureEditor (SFE) \cite{bobkov2024devil} focuses on detail-rich inversion by refining feature representations, enabling high-quality editing while preserving facial attributes. However, these methods still suffer from a fidelity-editability trade-off, limited generalization beyond the training domain, and high computational costs.

Recently, regression-based variants \cite{khan_mr_ijcv_2020, akram2021-pixel_fes, akram23lsrf} have been used for synthesizing realistic expressions. Khan \etal \cite{khan_mr_ijcv_2020} proposed the Masked Regression (MR) network, where each output pixel observes a local patch in the input image. Akram and Khan \cite{akram2021-pixel_fes} extended MR by proposing pixel-based ridge regression, where each output pixel attends to a single input pixel. 
Although these regression methods produce realistic expressions and generalize relatively well to out-of-distribution images, they often induce blurring artifacts. In this work, we aim to improve GAN generalization by combining a ridge regression-based loss with the adversarial loss.

The novelty of RegGAN lies in combining the strengths of regression- and GAN-based approaches. GAN-based methods fail to generalize on out-of-distribution faces, while regression methods generalize well but yield blurry results. RegGAN introduces a patch-based regression layer to ensure generalization and a spatial attention-based refinement network. The regression loss forces the network to learn accurate expressions and preserve facial details, while the adversarial loss enhances realism and visual fidelity. RegGAN generalizes well to out-of-distribution images, despite being trained on only a few hundred samples from the CFEE \cite{du-2014} dataset, which contains images captured in a controlled environment with consistent lighting and background.
This combination resolves both limitations and achieves robust expression synthesis across diverse facial images, including portraits, avatars, statues, and cartoons that come from distributions significantly different from the distribution of training images.

\begin{figure*}
    \centering
    \includegraphics[width=\linewidth]{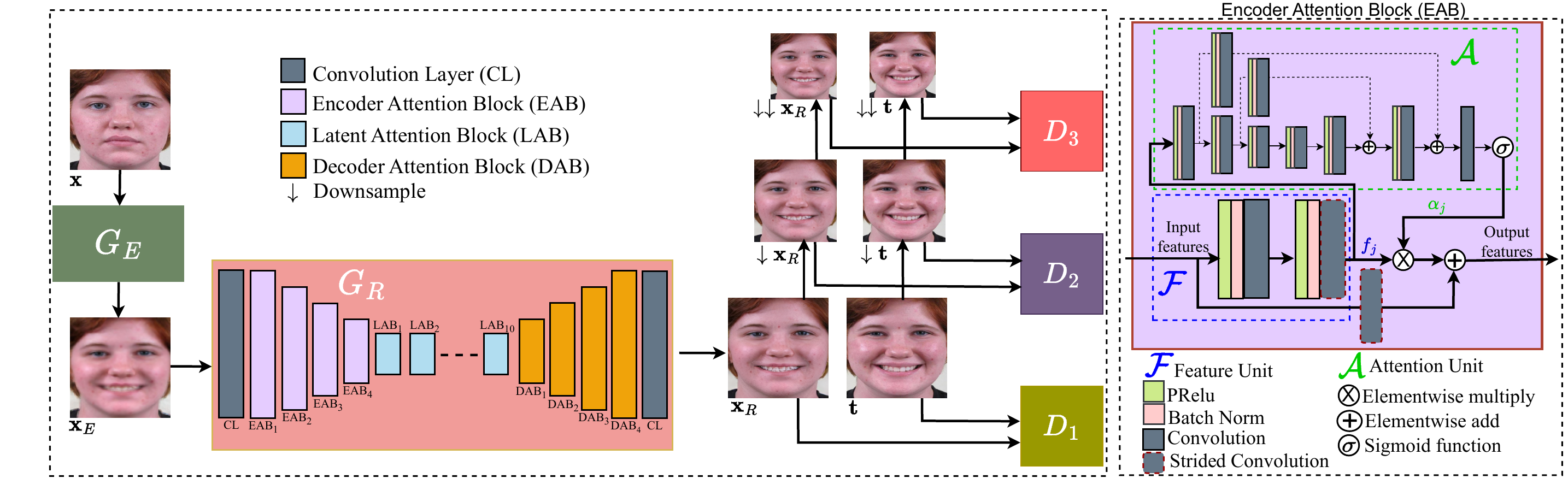}
    \caption{\textbf{Left:} Our method consists of two components: an expression layer and a refinement network. The expression layer takes an input image and generates a new image of the same person with a different facial expression. The refinement network then enhances the quality of the synthesized image, making it more realistic and sharper. \textbf{Right:} Architecture of EAB.}     
    \label{fig:RegGAN}
\end{figure*}

\section{Proposed Regression GAN (RegGAN)}
\label{sec:proposed_method}
In this section, we introduce our proposed RegGAN model. We first describe the GAN framework in general, and then explain the different components that make up our model.

\subsection{Generative Adversarial Networks}
\label{subsec:gans}
Generative Adversarial Networks (GANs) \cite{goodfellow2014} aim to learn the process of generating samples from a given data distribution. Rather than modeling the distribution explicitly, GANs learn to produce samples directly. The adversarial learning framework encourages the generated samples to lie closer to the manifold of real data.

A GAN consists of two competing networks: a generator $G$ and a discriminator $D$. The objective function for training GANs is defined as
\begin{align}
    \mathcal{L}(\theta_G,\theta_D)=&\mathbb{E}_{\mathbf{y}}\left[\log(D(\mathbf{y}))\right]+\mathbb{E}_{\mathbf{z}}\left[\log(1-D(G(\mathbf{z})))\right],
\end{align}
where $\mathbf{y}$ is a sample from the real data distribution, $\mathbf{z}$ is random noise, and $\theta_D$ and $\theta_G$ are the parameters of the discriminator $D$ and generator $G$, respectively. 
Conditioning $G$ and $D$ on input images and/or labels enables the modeling of conditional distributions \cite{mirza2014conditional}.
The objective function for conditional GANs, given conditioning input $\mathbf{y}$, is defined as
\begin{align}    
\mathcal{L}(\theta_G,\theta_D)=&\mathbb{E}_\mathbf{y}\left[\log(D(\mathbf{y},\mathbf{x}))\right]+ \\ \nonumber &\mathbb{E}_{\mathbf{z},\mathbf{x}}\left[\log\log(1-D(G(\mathbf{z},\mathbf{x}),\mathbf{x}))\right].
\end{align}
This is exploited in \cite{isola-2016} to perform image-to-image translation.

\begin{figure*}[t]
    \centering
    \renewcommand{\arraystretch}{0.5}
    \setlength{\tabcolsep}{0.1pt}
    \begin{tabular}{cccccc}
     $\mathbf{x}$ & $G_E(\mathbf{x})$ & $G_R(G_E(\mathbf{x}))$ &
     $\mathbf{x}$ & $G_E(\mathbf{x})$ & $G_R(G_E(\mathbf{x}))$ \\
     \includegraphics[width=.14\linewidth]{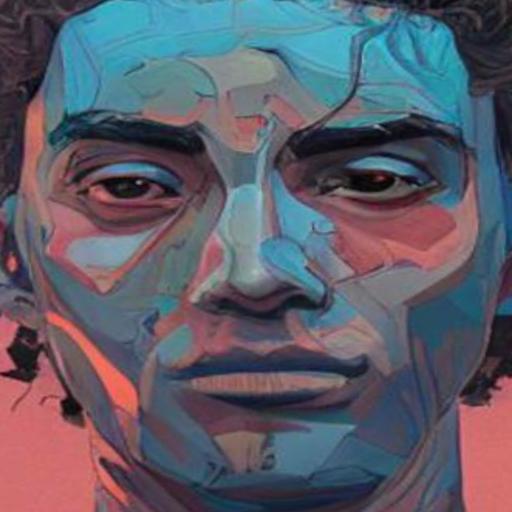} &
     \includegraphics[width=.14\linewidth]{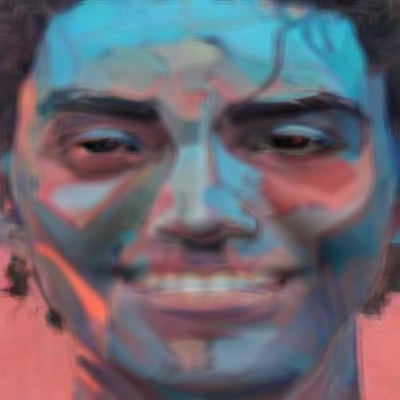} &
     \includegraphics[width=.14\linewidth]{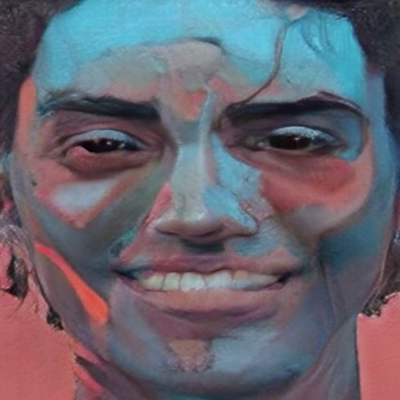} &
     \includegraphics[width=.14\linewidth]{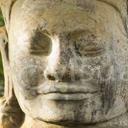} &
     \includegraphics[width=.14\linewidth]{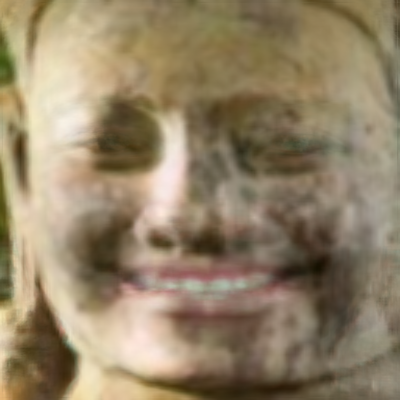} &
     \includegraphics[width=.14\linewidth]{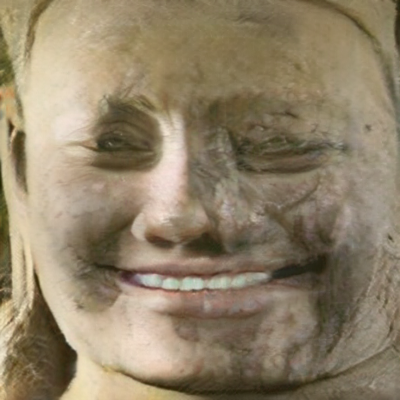} \\
    \end{tabular}
    \caption{Illustration of the input, intermediate, and final outputs synthesized by our proposed RegGAN. The expression layer $G_E$ generates an intermediate image that captures the target expression, while the refinement network $G_R$ transforms this intermediate result into a convincing photorealistic output.}
    \label{tab:impact_of_refinement}
\end{figure*}

\subsection{Regression GAN (RegGAN)}
\label{subsec:proposed_reggan}
We learn a regression-based expression translator and an adversarial-based refinement network within a joint generator-discriminator framework. The generator network can be divided into two parts: i) the expression layer $G_E$ and ii) the refinement network $G_R$. The expression layer $G_E$, trained with a sparsified regression objective, learns accurate expression biases and generates an intermediate image that reliably captures the target expression. This step facilitates generalization across a variety of images. The subsequent refinement network $G_R$ then converts this intermediate image into the final output, producing a realistic image while preserving the desired expression. Figure \ref{tab:impact_of_refinement} illustrates the roles of the expression layer and refinement network.

Mathematically, for an input image $\mathbf{x}$, the generator output can be represented as $\mathbf{x}_E=G(\mathbf{x})=G_R(G_E(\mathbf{x}))$, corresponding to a sequential translation of the input sample. The overall loss function is defined as the sum of sequential regression and adversarial losses

\begin{align}
    \mathcal{L}(\theta_{E},\theta_R,\theta_D) &= \mathcal{L}_E^\text{Reg}(\theta_{E})+\lambda \mathcal{L}_\text{GAN}(\theta_R, \theta_D) 
    \label{eq:loss}
\end{align}
where $\theta_E$, $\theta_R$, and $\theta_{D}$ are the parameters of the expression layer and refinement network, respectively. The details of the expression layer $G_E(\mathbf{x})$ are provided in Subsection~\ref{subsec:shallow_expression_mapping}, while the architectures of the GAN generator and discriminator, as well as their respective loss functions, are discussed in Subsections~\ref{subsec:refinement_network} and \ref{subsec:discrminator}.

Minimization of \eqref{eq:loss} causes the expression layer $G_E$ to be learned through supervised as well as adversarial learning, with $\lambda$ controlling their relative importance. Supervised learning ensures that $\theta_E^\ast$ generalizes well to out-of-distribution images, as shown in Subsection~\ref{subsec:shallow_expression_mapping}. After training $G_E$, its outputs are used as inputs for the refinement network $G_R$, which is then trained using adversarial learning to produce photo-realistic images.

\subsection{Expression layer}
\label{subsec:shallow_expression_mapping}
The expression layer acts as a shallow network that projects samples from the input distribution to the output distribution. This helps RegGAN to generalize well over out-of-dataset images. The expression layer weights can be obtained using masked regression \cite{khan_mr_ijcv_2020}, which is a sparsified modification of ridge regression. Specifically, each pixel is obtained as a linear combination of its local neighborhood using $r\times r$ receptive fields. This reduces the number of learnable parameters
for each pixel to $r^2$ compared to the $mn$ parameters used in ridge regression for the input image of size $m \times n$. Such sparse as well as localized receptive fields have been shown to help avoid over-fitting and improve out-of-distribution generalization \cite{khan_mr_ijcv_2020, akram2021-pixel_fes}. 
`
Let $\mathbf{x}_p^n\in\mathbb{R}^{r^2}$ represent the vectorized $r\times r$ input to the  $p^{\text{th}}$ unit for the  $n^{\text{th}}$ training sample. Let $t_p^n\in\mathbb{R}^1$ be the corresponding target value. For a training set of $N$ input and target pairs, we can construct the design matrix $\mathbf{X}_p\in\mathbb{R}^{N\times {r^2}}$ and the target vector $\mathbf{t}_p\in\mathbb{R}^{N\times 1}$ as
\begin{eqnarray}
    \mathbf{X}_p =
    \begin{bmatrix}
    \mathbf{x}_p^{1\top} \\ \mathbf{x}_p^{2\top} \\ \vdots \\ \mathbf{x}_p^{N\top}
    \end{bmatrix}
& \text{and} &
    \mathbf{t}_p =
    \begin{bmatrix}
    t_p^1 \\ t_p^2 \\ \vdots \\ t_p^N
    \end{bmatrix}
    \label{eq:X_pt_p}
\end{eqnarray}
The loss function for the receptive field weights $\mathbf{w}_p\in\mathbb{R}^{r^2\times1}$ and bias $b_p\in\mathbb{R}^1$ for the $p^{\text{th}}$ pixel can be written as
\begin{align}
    \mathcal{L}_E^\text{Reg}(\mathbf{w}_p,b_p) &= \left\Vert \mathbf{X}_p\mathbf{w}_p + b_p\mathbf{1} - \mathbf{t}_p \right\Vert_2^2 + \frac{\lambda_\text{Reg}}{2}\left(\left\Vert\mathbf{w}_p\right\Vert_2^2+b_p^2\right)
    \label{eq:G_E_loss}
\end{align}
where $\mathbf{1}\in\{1\}^N$ is a vector containing all ones. Parameters that globally minimize this quadratic loss function can be uniquely determined in closed-form as
\begin{align}
    \begin{bmatrix}
    \mathbf{w}_p\\b_p
    \end{bmatrix}
    &=
    \begin{bmatrix}
    \mathbf{X}_p^\top\mathbf{X}_p+\lambda_\text{Reg}\mathbf{I} & \mathbf{X}_p^\top\mathbf{1}
    \\
    \mathbf{1}^\top\mathbf{X}_p & \lambda_\text{Reg}+N
    \end{bmatrix}^{-1}
    \begin{bmatrix}
    \mathbf{X}_p^\top\mathbf{t}_p
    \\
    \mathbf{1}^\top\mathbf{t}_p
    \end{bmatrix}
    \label{eq:mr_system}
\end{align}

For receptive fields of size $r\times r$, the linear system \eqref{eq:mr_system} has an $(r^2+1)\times (r^2+1)$ system matrix which is quite small compared to the case of global receptive fields of ridge regression. For multi-channel/color images, we can learn separate regression mappings for each channel via repeated applications of \eqref{eq:mr_system}.

We observe that the training of the expression layer is quite efficient. For each image-to-image translation task, a different expression layer needs to be learned. For the case of expression mappings, for each expression pair we train a different expression layer. Since the expression layer projects the input to an intermediate representation, the rest of the network does not need to be retrained each time. Rather, it projects that intermediate representation to the output distribution independent of the distribution of the input data. This is the key to obtaining generalization across input distributions as well as across multiple image-to-image translation tasks.

\subsection{Refinement Network}
\label{subsec:refinement_network}
As shown in \autoref{fig:RegGAN}, RegGAN consists of two modules: a regression-based expression layer and an encoder-decoder-based refinement network. The refinement network takes the intermediate representation $\mathbf{x}_E$, passes this representation through an encoder-decoder network, and generates the output image $\mathbf{x}_R$.

The refinement network employs attention blocks to recover facial details of $\mathbf{x}_E$ at multiple scales. It consists of three encoding attention blocks (EABs), ten latent attention blocks (LABs) in the bottleneck, and three decoding attention blocks (DABs). 

Each attention block consists of a feature unit $\mathcal{F}$ and an attention unit $\mathcal{A}$. The feature unit extracts facial features, while the attention unit focuses on informative regions such as the eyes and mouth.

The attention unit employs an hourglass network \cite{newell2016stacked} followed by a convolutional layer to produce an attention map $\bm \alpha$. The hourglass network is designed to capture spatial dependencies among facial regions by combining bottom-up and top-down processing approach at multiple scales. This approach has proven useful in face super-resolution \cite{kim2021edge, yu2018face} and face alignment \cite{bulat2017far}. While the attention unit is shared across all block types  (EAB, LAB, DAB), the feature unit varies.

EAB employs two pre-activation convolutional blocks (PCBs) with PReLU \cite{he2015delving} activation functions to extract facial features. The second block uses strided convolution to downsample the feature volume, capturing a larger context while reducing computational cost. In contrast, DAB applies nearest-neighbor upsampling to enlarge the feature volume before passing it to the convolutional blocks.

The input feature of the $j^{th}$ EAB can be denoted as EAB$_{j-1} \in \mathbb{R}^{C_{j-1} \times H_{j-1} \times W_{j-1}}$. The outputs of $\mathcal{F}$ and $\mathcal{A}$ are computed using the following equations:
\begin{align}
    \mathbf{f}_j &= \mathcal{F}(EAB_{j-1}), \\
    \bm \alpha_j &= \sigma(\mathcal{A}(\mathbf{f}_j)).
\end{align}
Here $\mathbf{f}_j$ denotes the output of the feature unit $\mathcal{F}$, $\bm \alpha_j$ represents the output of the attention unit $\mathcal{A}$, and $\sigma$ is the sigmoid activation function. The output of EAB$_{j-1}$ can be written as
\begin{align}
    EAB_j = \Phi(EAB_{j-1}) + \bm\alpha_j \otimes \mathbf{f}_j,
\end{align}
where $\Phi(\cdot)$ is the strided convolution layer and $\otimes$ is  element wise multiplication operation.  

We use a combination of loss functions to train the refinement network. Similar to \cite{chen2020learning}, we adopt four losses: pixel loss $\mathcal{L}_{\text{pix}}$, adversarial loss $\mathcal{L}_{\text{adv}}$, feature matching loss $\mathcal{L}_{\text{fm}}$, and perceptual loss $\mathcal{L}_{\text{per}}$.
The overall objective is a weighted sum of these four terms, defined as
\begin{align}
    \mathcal{L}_R &= \lambda_{\text{pix}} \mathcal{L}_{\text{pix}} + \lambda_{\text{adv}} \mathcal{L}_{\text{adv}}^G + \lambda_{\text{fm}} \mathcal{L}_{\text{fm}} + \lambda_{\text{per}} \mathcal{L}_{\text{per}}.
    \label{eq:full_loss}
\end{align}

\noindent \textbf{Pixel Loss:} The pixel loss can be computed as 
\begin{align}
    \mathcal{L}_{\text{pix}} = \frac{1}{m} \sum_{n=1}^m \Vert G_R \left(\mathbf{x}_{E_n}\right) - \mathbf{t}_n \Vert_1,
    \label{eq:pixel_loss}
\end{align}
where $m$ is the number of training samples in the batch, $G_R$ represents the refinement network generator, $\mathbf{x}_{E_n}$ is the intermediate representation of the $n^{\text{th}}$ input image, $\mathbf{t}_n$ is the $n^{\text{th}}$ target image, and $\Vert . \Vert_1$ denotes the $\ell_1$ norm. By minimizing this loss, the refinement network learns to generate high-quality facial images that closely resemble the target images.

\noindent\textbf{Adversarial Loss:} To encourage the generator to produce photo-realistic images with sharp details, we adopt an adversarial loss on the generator. It is defined as:
\begin{align}
    \mathcal{L}_{\text{adv}}^G = \frac{1}{m} \sum_{n=1}^m \sum_{k=1}^3 - D_k \left(G_R (\mathbf{x}_{E_n})\right),
\end{align}
where $D_k$ represents the discriminator at scale $k$. This loss drives the generator outputs closer to real images in terms of perceptual quality.

\noindent\textbf{Feature Matching Loss:} To stabilize GAN training and improve the quality of the generated images, we adopt the feature matching loss \cite{wang2018high}. This loss minimizes the difference between the feature maps of the intermediate representation $\mathbf{x}_{E}$ and high-quality target images $\mathbf{t}$. It is computed as
\begin{align}
    \mathcal{L}_{\text{fm}} = \frac{1}{m} \sum_{n=1}^m \sum_{k=1}^3 \sum_{i=1}^j   \frac{1}{s_k^i} \Vert f_{D_k}^i(G_R (\mathbf{x}_{E_n})) - 
    f_{D_k}^i( \mathbf{t}_n) \Vert_1,
    \label{eq:feature_matching}
\end{align}
where $j$ is the total number of layers in $D_k$, $f_{D_k}^i$ denotes the feature maps at the $i^{\text{th}}$ layer of $D_k$, and $s_k^i$ is the total number of elements in  $f_{D_k}^i$.

\noindent\textbf{Perceptual Loss:} The perceptual loss measures the difference in perceptual quality between the features of the intermediate representation $\mathbf{x}_E$ and the ground-truth $\mathbf{t}$ images, using a pre-trained VGG-19 network \cite{simonyan2014very}. Following the notation in \eqref{eq:feature_matching}, the perceptual loss is defined as
\begin{align}
    \mathcal{L}_{\text{per}} = \frac{1}{m} \sum_{n=1}^m \sum_{i=1}^{l} 
    \frac{1}{s^i} \Vert  f_{\text{vgg}}^i  (G_R (\mathbf{x}_{E_n})) -  
     f_{\text{vgg}}^i( \mathbf{t}_n) \Vert_1.
    \label{eq:perceptual_loss}
\end{align}
where $l$ is the total number of VGG-19 layers used for computing the perceptual loss, $f_{\text{vgg}}^i$ denotes the feature maps at the $i^{\text{th}}$ layer of VGG-19, and $s^i$ is the number of elements in $f_{\text{vgg}}^i$.

\subsection{Discriminator}
\label{subsec:discrminator}
We employed 3 discriminators \cite{wang2018high}, denoted as $D_1, D_2$ and $D_3$, with identical architectures in the refinement network. These discriminators are designed to capture fine-grained facial details as well as global and local structures at multiple scales. This design enables the refinement network to generate high-quality facial images that might be missed when using a single discriminator. The effectiveness of multiple discriminators has been demonstrated in facial image manipulation \cite{lee2020maskgan, nirkin2019fsgan} and face super-resolution \cite{chen2020learning, chen2021progressive}. 

We construct a three-scale image pyramid by downsampling the output images $\mathbf{x}_R$ and the target image $\mathbf{t}$ by factors of $2$ and $4$, respectively. The hinge-based loss \cite{lim2017geometric}, commonly used for multi-scale discriminators, is defined as 
\begin{align}
    \mathcal{L}_{\text{adv}}^D  =  \frac{1}{m} \sum_{n=1}^m \sum_{k=1}^3 [\max(0, 1 - D_k(\mathbf{t}_n) +  \\ \nonumber
    \max(0, 1 + D_k(\mathbf{x}_{E_n})].
\end{align}

\begin{figure*}[t]
    \centering
    \includegraphics[width=\linewidth]{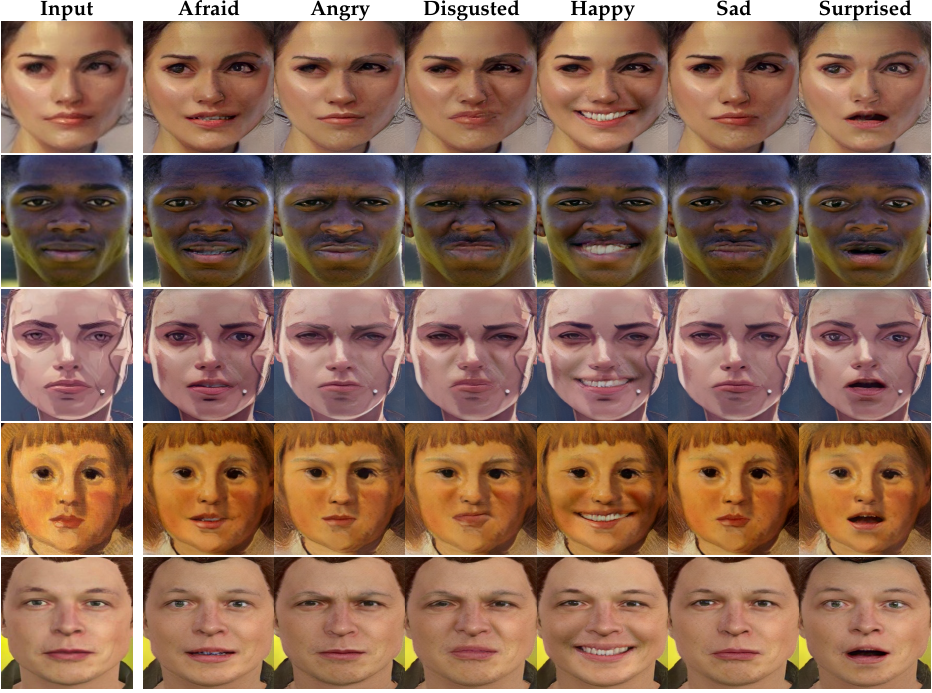}
    \caption{Results of RegGAN for facial expression synthesis on out-of-distribution images, including an impasto face (row 1), a celebrity face (row 2), a fantasy image (row 3), a portrait (row 4), as well as an avatar (row 5). Impasto and fantasy images were generated using stable diffusion \cite{stable}. The proposed method introduces convincing expressions while preserving the identity and facial details of the input image.}
    \label{fig:out-of-dist-celeb}
\end{figure*}

\section{Experimental Evaluation}
\label{sec:exp_evaluation}
\subsection{Datasets}
The expression layer and the refinement network are trained using the subset of the Compound Facial Expressions of Emotion (CFEE)\cite{du2014compound} database, containing only the seven universally recognized facial expressions - afraid, angry, disgusted, neutral, happy, sad, and surprised. Specifically, we have $230$ images per expression, resulting in a total of $1,610$ images. We used a $90$/$10$ split for training and testing.

The refinement network was initially trained on FFHQ \cite{karras2019style} dataset, which contains $70,000$ high-quality facial images. All facial images were aligned, cropped, and then resized to $400 \times 400$.

For the results of StarGAN \cite{choi-2017} and MR \cite{khan_mr_ijcv_2020}, we trained the models on the CFEE dataset. The results for GANimation  \cite{pumarola2018ganimation}, SARGAN \cite{akram23sargan}, US-GAN \cite{akram23usgan}, and DAI2I \cite{chen2020domain} were obtained using pre-trained models.

For quantitative evaluation, we used $22$ images from CFEE dataset and $22$ additional images of celebrities, portraits, and avatars collected from publicly available sources on the Internet. These images can be accessed at \footnote{\url{https://github.com/arbishakram/RegGAN/tree/main/testing-imgs}}. 

\subsection{Implementation Details}
The proposed method is implemented using the PyTorch framework. The Adam optimizer \cite{kingma2014adam} is employed for training $G_R$, $D_1$, $D_2$, and $D_3$ with $\beta_1=0.5$, $\beta_2=0.999$, and batch size of $1$. Following \cite{chen2020learning}, we set the loss weights in \eqref{eq:full_loss} to $\lambda_{\text{pix}}=100$, $\lambda_{\text{adv}}=1$, $\lambda_{\text{fm}}=10$, and $\lambda_{\text{per}}=1$. The value of $r$ is set to $5$ for all experiments.

\begin{figure*}[t!]
    \centering
    \includegraphics[width=\linewidth]{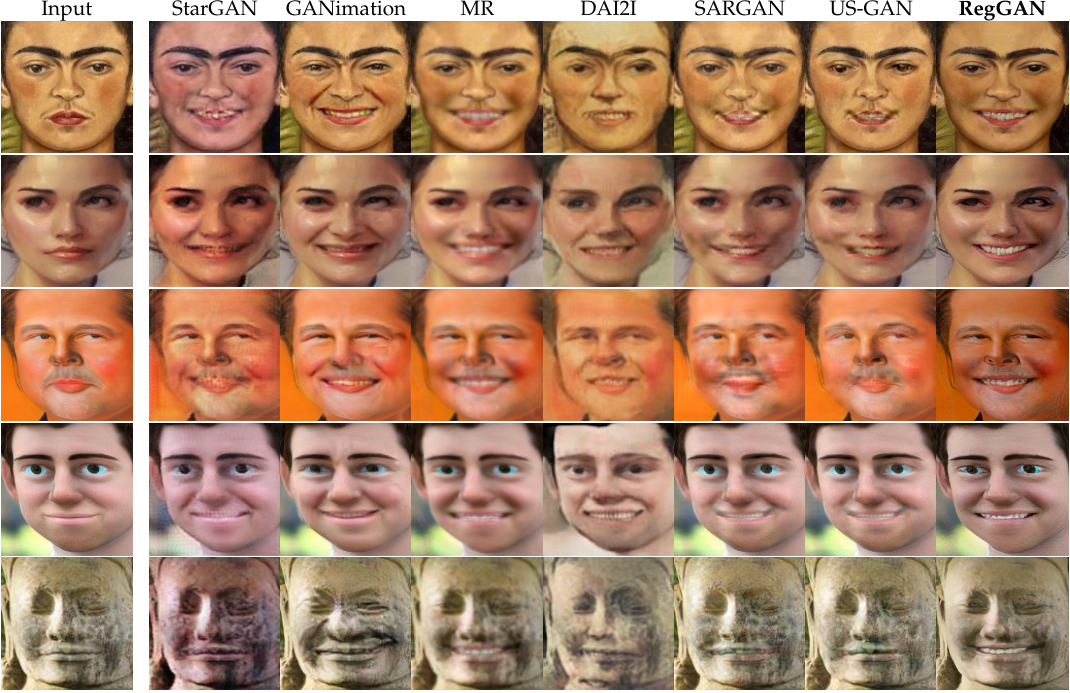}
    \caption{Comparison of the proposed method, RegGAN, with six state-of-the-art facial expression synthesis models - StarGAN, GANimation, MR, DAI2I, SARGAN as well as US-GAN on out-of-distribution facial images. GANimation synthesizes realistic expression but introduces noticeable artifacts. StarGAN and DAI2I fail to preserve the input image's color distribution in their output. SARGAN and US-GAN preserve facial details but are unable to induce happy expressions. MR generates realistic happy expressions, but the results are often blurry. In contrast, RegGAN produces sharper and more realistic expressions on out-of-distribution images. Moreover, RegGAN demonstrates consistent performance across diverse image types, including human portraits, status, and avatars collected from the Internet, despite being trained exclusively on CFEE dataset captured in a controlled environment with consistent lighting and background.}    
    \label{fig:comp_with_state_of_the_art_happy_exp}
\end{figure*}

\subsection{Color Data Augmentation}
The training of RegGAN relies on paired data. Specifically, the regression-based expression layer is trained using neutral-target expression image pairs, while the refinement network is trained on the output of the expression layer paired with the corresponding ground-truth target expression.
Supervised models often require a large number of training samples to achieve satisfactory results. To further increase the diversity of the training dataset and enhance the performance of the expression layer, we apply color augmentation to the neutral–target expression image pairs. This technique preserves the shape and geometry of the original face image while changing colors to simulate multiple skin tones. Color augmentation increases the diversity of samples with different skin tones, which enhances the generalization performance of the regression layer. We use a pre-trained model, ReHistoGAN \cite{afifi2021histogan}, to recolor images. This model takes an input image and a target image to generate an output image with the color distribution of the target. For each image, we create ten augmented versions. Additional details of the augmentation process are provided in the supplementary material.

\subsection{Evaluation Metrics}
We quantitatively evaluated the quality of synthesized images using the four objective metrics: i) Expression Classification Score, ii) Face Similarity Score, iii) Realism Score, and iv) Fr\'echet Inception Distance, as well as a user study for subjective evaluation.
\subsubsection{Expression Classification Score (ECS)}
We trained a ResNet-18 \cite{he2016deep} on the CFEE dataset for facial expression recognition. This classifier was then used to evaluate whether the target expressions were correctly synthesized. ECS is calculated as the number of correctly classified images divided by the total number of images.
\subsubsection{Face Similarity Score (FSS)}
To assess whether the proposed method preserves facial details, we employ the face similarity score (FSS). We compute the similarity between the synthesized expression images and the input images using OpenFace \cite{amos2016openface}.

\subsubsection{Realism Score (RS)}
To measure the quality of synthesized images, we used the QualiCLIP \cite{qualiclip2024} score. The CLIP encoder \cite{radford2021learning} extracts embeddings for both synthesized images and seven positive-negative prompt pairs.  
The prompt pairs are: (`Good photo', `Bad photo'), (`Sharp image', `Blurry image'), (`Sharp edges', `Blurry edges'), (`High-resolution image', `Low-resolution image'), (`Noise-free image', `Noisy image'), (`High-quality image', `Low-quality image'), and (`Good picture', `Bad picture'). For each pair, cosine similarities are converted to probabilities with a softmax, and the probability of the positive prompt is selected. The final score is the mean of these probabilities across all pairs. 

\subsubsection{Fr\'echet Inception Distance (FID)}
To evaluate how well the generated images approximate the real data distribution, we used the FID \cite{heusel2017gans}. A pretrained Inception network \cite{szegedy2016rethinking} is employed to extract the features of synthesized and real expression images. Each feature set is represented by its mean and covariance, forming a multivariate Gaussian. The FID is then computed as the distance between these two distributions, where a lower distance indicates that synthesized images are closer to the real images.

\subsubsection{User Study}
For subjective evaluation, ten images are randomly selected, and synthesized versions are generated using the proposed method and six state-of-the-art baselines: StarGAN, GANimation, MR, DAI2I, SARGAN, and US-GAN. These images can be found at this link \footnote{\url{https://github.com/arbishakram/RegGAN/tree/main/userstudy-fig}}. For each input image, seven happy-expression images are displayed, and participants are asked:
i) Which image appears the most realistic?
ii) Which image has the most convincing expression?
iii) Which image best maintains the identity of the original input?
Images were randomly shuffled, and the source models were not revealed.

\begin{figure*}[t]
    \centering
    \includegraphics[width=\linewidth]{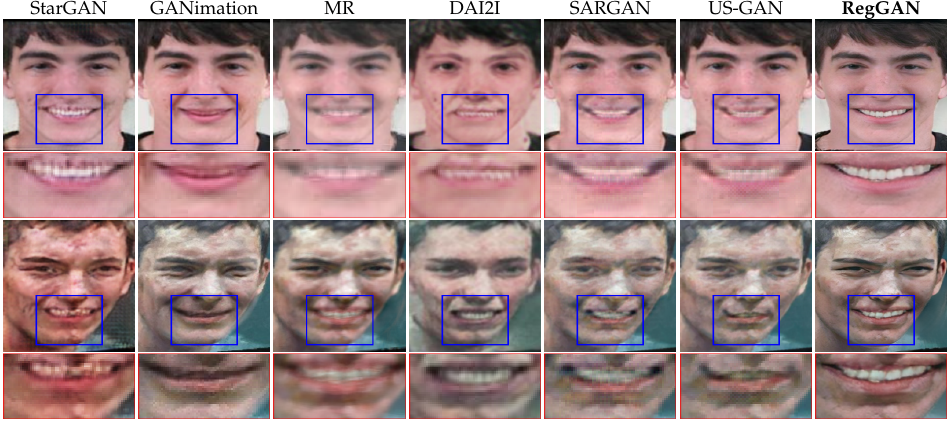}
    \caption{Zoomed-in versions of smiles from our proposed method as well as other methods. Our method generates more realistic smiles and enhances the clarity of individual teeth, improving their differentiation and visibility compared to all other methods.}
    \label{fig:reggan_comparison4}
\end{figure*}

\subsection{Qualitative Analysis}
Figure~\ref{fig:teaser_fig} demonstrates that RegGAN produces photo-realistic expressions and exhibits strong generalization to unseen images.
Figure~\ref{fig:out-of-dist-celeb} demonstrates the results of RegGAN on six universal expressions. For each input-target expression pair, a separate regression layer is trained. However, the rest of the network is trained only once. This experiment shows that RegGAN can quickly generalize to multiple expression-mapping tasks. Although it is trained exclusively on human expression datasets, RegGAN generalizes remarkably well to portraits, avatars, and paintings.

A comparison of the proposed RegGAN with five state-of-the-art methods is presented in \autoref{fig:comp_with_state_of_the_art_happy_exp}. The test set includes portraits (rows 1 and 3), an impasto face (row 2), a cartoon face (row 4), and a statue (last row). 
The results show that GANimation \cite{pumarola2018ganimation} synthesizes realistic facial expressions but introduces noticeable artifacts. It is trained on $200,000$ images, while our method uses $5000$ images after applying data augmentation. GANimation requires precise facial action unit extraction, whereas
our method operates without it. US-GAN \cite{akram23usgan} and SARGAN fail to generate convincing happy expressions. In contrast, StarGAN \cite{choi-2017} and DAI2I \cite{chen2020domain} struggle to preserve the color and identity details of the input images. The regression-based method, MR \cite{khan_mr_ijcv_2020}, generalizes better than the GAN-based approaches but produces blurry results. In contrast, the proposed RegGAN generates sharper and more convincing expressions on out-of-distribution images. These results demonstrate that, despite being trained on a smaller expression synthesis dataset, RegGAN achieves superior performance.

\autoref{fig:reggan_comparison4} presents a zoomed-in comparison of smiles generated by RegGAN and competing methods. Existing methods often produce blurry smiles, lack precise lip contours, and fail to capture the natural structure of the teeth. They typically add whiteness to the mouth region without capturing finer details. In contrast, RegGAN generates realistic smiles with enhanced detail, preserving the natural curvature of the lips and capturing the fine structure of teeth for clearer differentiation.

\begin{table}[t]
    \centering
    \begin{tabular}{lcccc}   
       Method  & ECS $\uparrow$ & FSS $\uparrow$ & RS $\uparrow$ & FID $\downarrow$ \\ \hline
       StarGAN \cite{choi-2017}  & 0.71  & 0.68 & 0.55 & 111.9 \\
       GANimation \cite{pumarola2018ganimation} & 0.45 & 0.64 & 0.80 & 94.09 \\
       MR \cite{khan_mr_ijcv_2020} & 0.72  & 0.66 & 0.19 & 113.9 \\
       DAI2I \cite{chen2020domain} & 0.54 & 0.64 & 0.58 & 140.2 \\
       SARGAN \cite{akram23sargan} & 0.61  & \textbf{0.71} & 0.73 & 98.86 \\
       US-GAN \cite{akram23usgan} & 0.66  & 0.71 & 0.69 & 98.92 \\
       \textbf{RegGAN}  & \textbf{0.77}  & 0.69 & \textbf{0.84} & \textbf{72.48} \\  \hline
    \end{tabular}
    \caption{RegGAN outperforms other methods in ECS and RS by synthesizing convincing, photo-realistic facial expressions. The lowest FID indicates that the generated images closely match the distribution of real images. In FSS, it attains the second-best value, as SARGAN and US-GAN use ultimate skip connections that directly copy input details to the output. 
    }
    \label{tab:quant-analysis}
\end{table}

\begin{figure}[t]
    \centering
    \includegraphics[width=\linewidth]{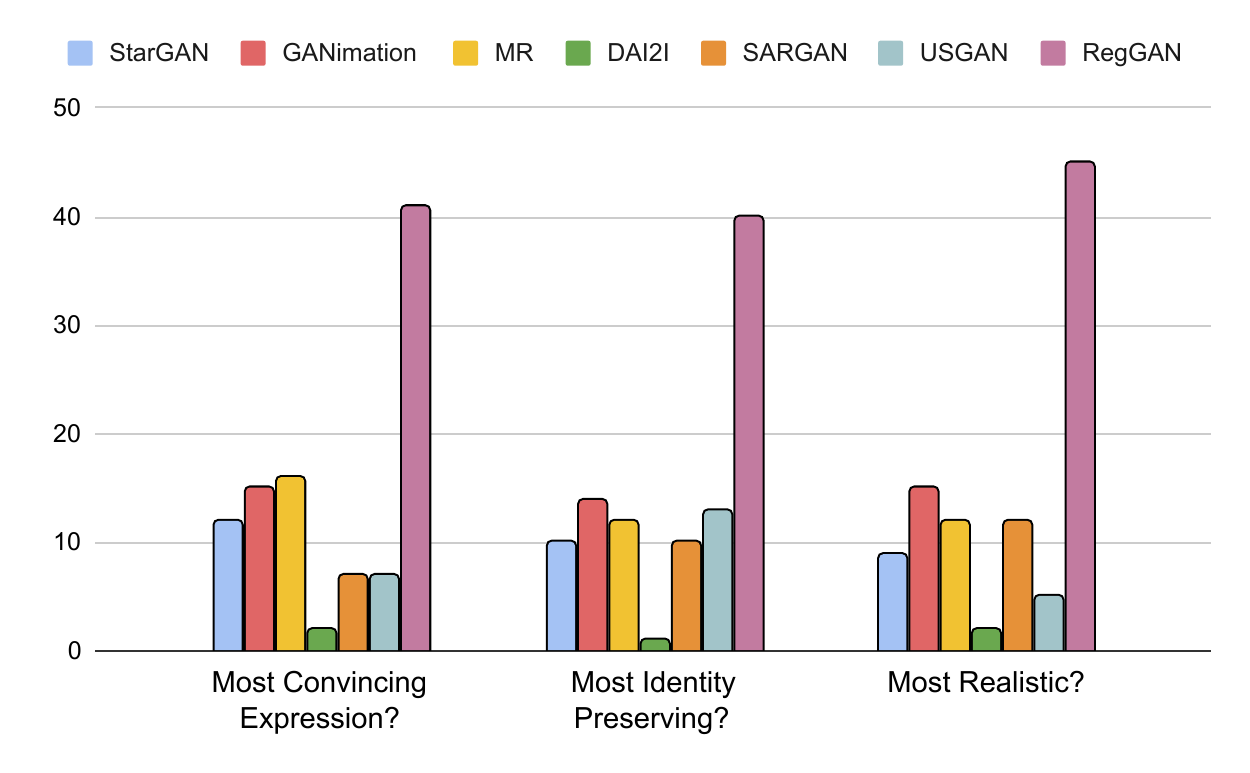}
    \caption{RegGAN outperforms all existing baselines in synthesizing convincing and realistic expression, while preserving facial details.}
    \label{fig:userstudy_results}
\end{figure}

\subsection{Quantitative Analysis}
\label{sec:quantiative-analysis}
In Table~\ref{tab:quant-analysis}, we present a quantitative comparison of RegGAN with six state-of-the-art methods. The regression-based method, MR, generates reasonably convincing expressions while preserving facial details. However, its performance drops in RS and FID. While SARGAN and US-GAN maintain facial details (FSS) via ultimate skip connections, they struggle to produce convincing expressions. In contrast, RegGAN balances expression synthesis, identity preservation, and realism better than all other methods.
Figure~\ref{fig:userstudy_results} summarizes the results of a user study involving forty participants. These results corroborate the objective evaluation metrics, showing that RegGAN outperforms all other methods in expression, identity preservation, and realism by receiving the highest votes.

\subsection{Ablation Studies}
\label{sec:ablation-studies}
To assess the impact of latent attention blocks and spatial attention, we conducted two ablations: 
\begin{enumerate}
    \item No Latent Attention Blocks (LABs): removing all latent attention blocks from the refinement network.
    \item No Attention Units (AUs): removing the spatial attention unit from each attention block.
\end{enumerate}
Experiments were conducted on $256 \times 256$ size images with batch size $3$. The refinement network was initially trained on the FFHQ dataset for $4$ epochs, followed by fine-tuning on the augmented CFEE dataset. For a fair comparison, we fine-tune RegGAN on $256 \times 256$ size images using the same training details.

\autoref{tab:quant-analysis-ablation} presents the quantitative results of these ablation studies. Including latent attention blocks and spatial attention units improves facial and expression detail preservation. Removing LABs increases ECS from $0.85$ to $0.87$ but raises FID to $80.8$, indicating a larger distribution shift. Removing AUs results in the worst FID and RS. The full model achieves the lowest FID with balanced performance across all metrics.

\begin{figure}[t]
  \centering
  \fontsize{9}{9}\selectfont
  \renewcommand{\arraystretch}{1}
  \setlength{\tabcolsep}{0.3pt}
     \begin{tabular}{cccc}
     Input & StyleRes \cite{pehlivan2023styleres} & SFE \cite{bobkov2024devil} &  RegGAN \\
     \includegraphics[width=.25\linewidth]{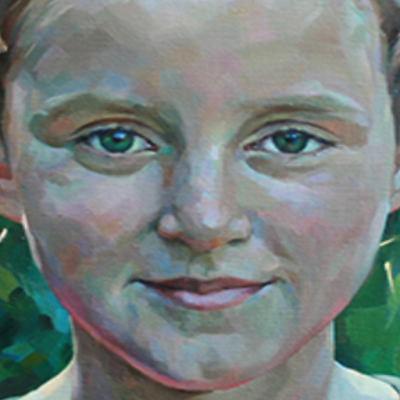}  & 
     \includegraphics[width=.25\linewidth]{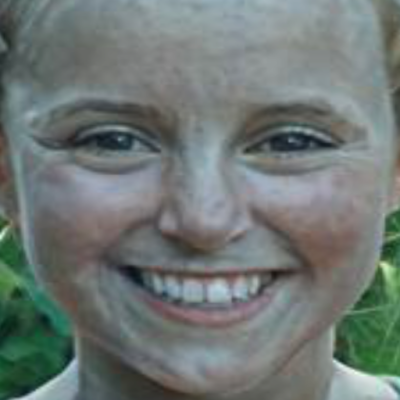}  &
     \includegraphics[width=.25\linewidth]{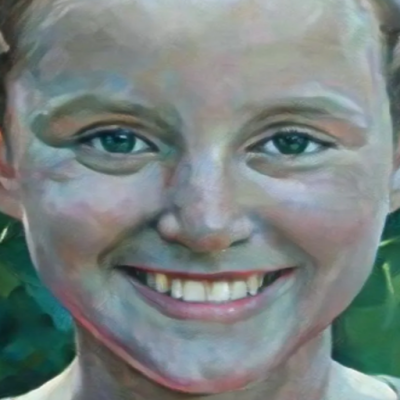}  &
     \includegraphics[width=.25\linewidth]{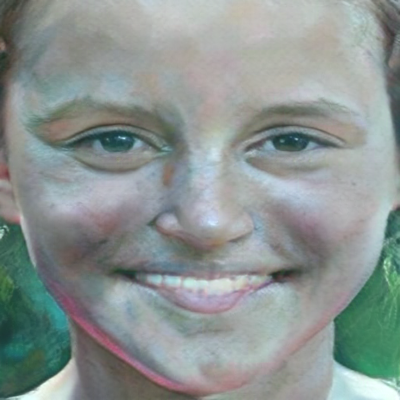} \\

     \includegraphics[width=.25\linewidth]{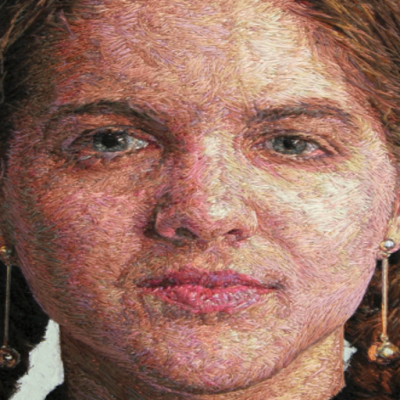}  & 
     \includegraphics[width=.25\linewidth]{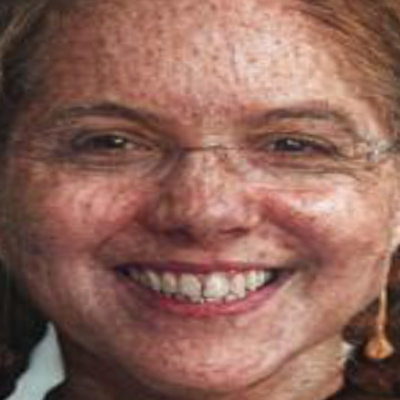}  &
     \includegraphics[width=.25\linewidth]{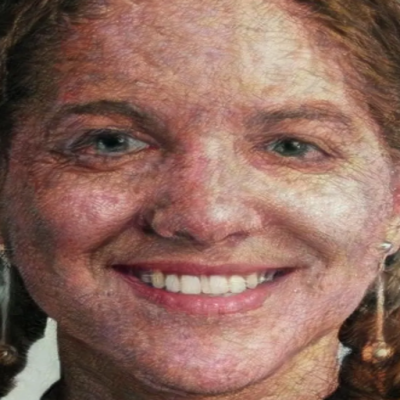}  &
     \includegraphics[width=.25\linewidth]{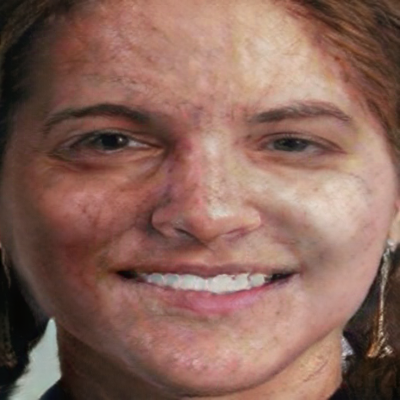} \\

     \includegraphics[width=.25\linewidth]{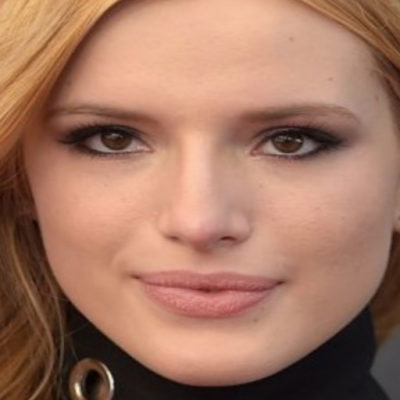}  & 
     \includegraphics[width=.25\linewidth]{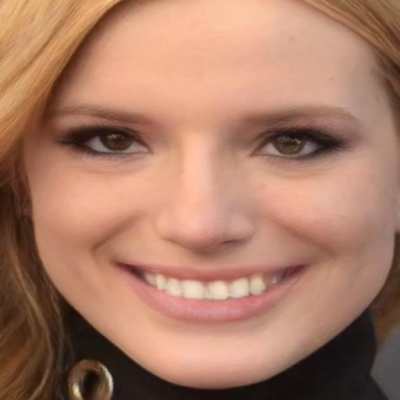}  &
     \includegraphics[width=.25\linewidth]{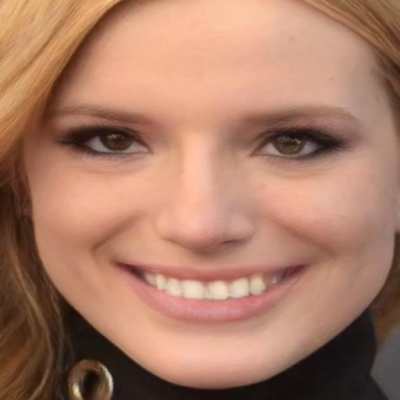}  &
     \includegraphics[width=.25\linewidth]{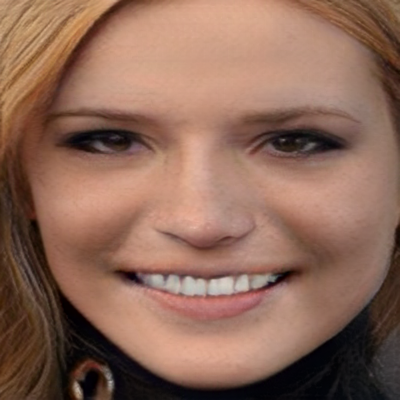} \\    
  \end{tabular}
  \caption{StyleGAN-based methods synthesize smiles but tend to lose fine facial details and introduce unwanted artifacts, such as glasses (third row, second column). The proposed approach better preserves identity details while synthesizing realistic facial expressions.}
  \label{fig:stylegan-comparison}
\end{figure}

\begin{table}[t]
    \centering    
    \begin{tabular}{lcccc}   
       Method  & ECS $\uparrow$ & FSS $\uparrow$ & RS $\uparrow$ & FID $\downarrow$ \\ \hline
       No LABs & \textbf{0.87} & 0.67 & 0.90 & 80.8 \\
       No AUs & 0.86 & 0.68 & 0.88 & 89.9 \\
       \textbf{RegGAN} & 0.85 & \textbf{0.68} & \textbf{0.90} & \textbf{74.3} \\ \hline
    \end{tabular}
    \caption{Quantitative results of ablation studies. The inclusion of LABs and AUs helps achieve a better balance among expression, facial details, and realism.}
    \label{tab:quant-analysis-ablation}
\end{table}

\begin{table}[t]
    \centering  
    \begin{tabular}{lcccc}   
       Method  & ECS $\uparrow$ & FSS $\uparrow$ & RS $\uparrow$ & FID $\downarrow$ \\ \hline        
        StyleRES \cite{pehlivan2023styleres} & 0.91 & 0.58 & 0.28 & 83.5 \\
        SFE \cite{bobkov2024devil} &  0.70 & 0.64 & 0.39 & 67.4 \\
        \textbf{RegGAN} & \textbf{0.93}  & \textbf{0.64} & \textbf{0.85} & \textbf{61.8} \\ \hline
    \end{tabular}
    \caption{Quantitative results of RegGAN and StyleGAN-based methods. RegGAN achieves higher scores in expression quality, identity preservation, and realism.}
    \label{tab:quant-analysis-stylegan}
\end{table}

\subsection{Comparison with StyleGAN variants}
In Figure~\ref{fig:stylegan-comparison} and Table~\ref{tab:quant-analysis-stylegan}, we compare the proposed method with StyleGAN-based variants both qualitatively and quantitatively. While StyleGAN-based methods can generate realistic facial expressions, they struggle to balance expression quality with fidelity. Their latent manipulations entangle expression with identity, leading to missing fine details and artifacts that degrade realism.
 
\section{Discussion}
\label{sec:discussion}
\emph{Limitations of Skip-Connection Based Methods:}
Since facial expressions do not affect facial details, SARGAN \cite{akram23sargan} and US-GAN \cite{akram23usgan} utilize ultimate skip connections for facial expression synthesis. These methods exhibit impressive results on out-of-distribution facial images by effectively transferring details from the input to the output. However, despite their success in preserving facial and color details, they struggle to generate convincing expressions. This limitation arises because they directly add the input image to the output without incorporating any scaling coefficient. In contrast, the proposed method, RegGAN, not only preserves the facial and color details but also induces convincing expressions as demonstrated in Figures~\ref{fig:out-of-dist-celeb}, \ref{fig:comp_with_state_of_the_art_happy_exp}, and \ref{fig:reggan_comparison4}.

\begin{figure}[t]
  \centering
  \fontsize{9}{9}\selectfont
  \renewcommand{\arraystretch}{1}
  \setlength{\tabcolsep}{0pt}
  \begin{tabular}{cccccc}
     Input & Disgusted & Surprised & Input & Angry & Surprised\\    
     \includegraphics[width=.15\linewidth]{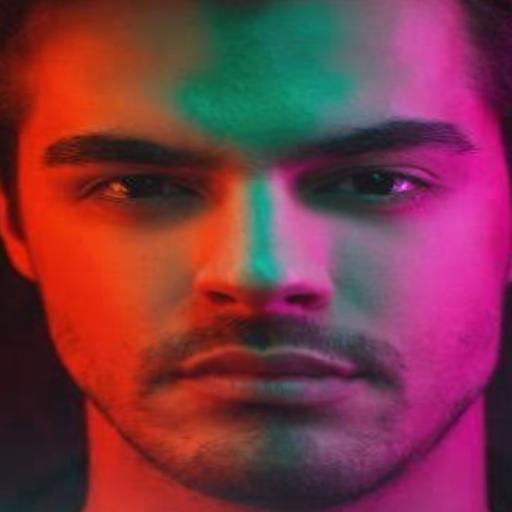}  & 
     \includegraphics[width=.15\linewidth]{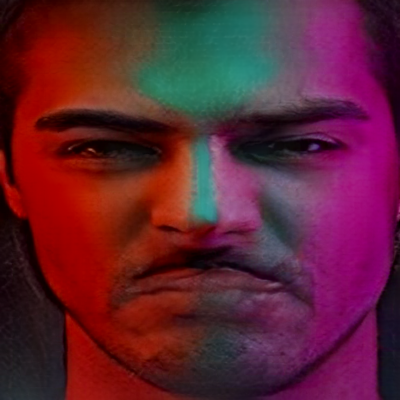} &
     \includegraphics[width=.15\linewidth]{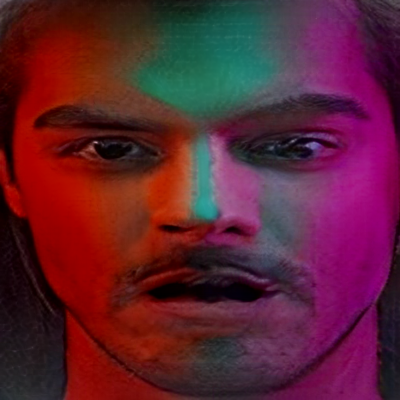} & \hspace{0.1cm}
     \includegraphics[width=.15\linewidth]{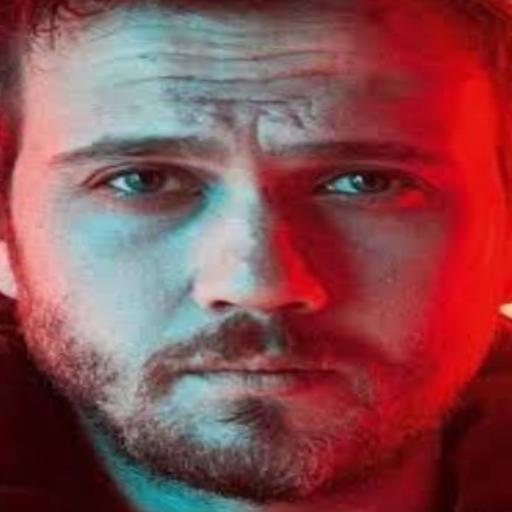} &
     \includegraphics[width=.15\linewidth]{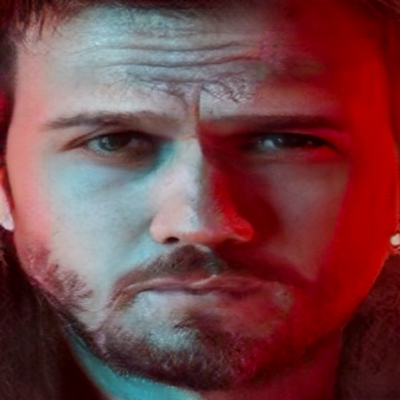} & 
     \includegraphics[width=.15\linewidth]{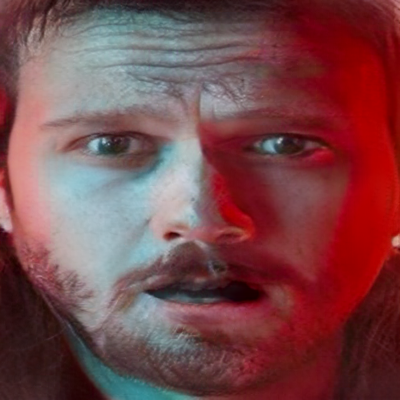}  \\

     \includegraphics[width=.15\linewidth]{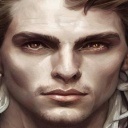}  & 
     \includegraphics[width=.15\linewidth]{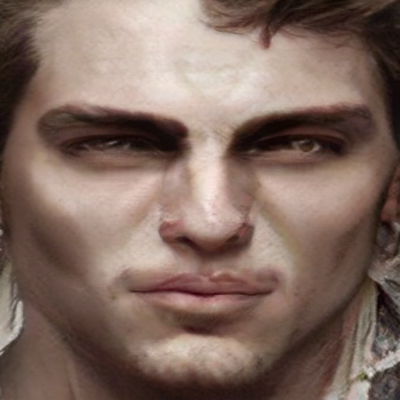} &
     \includegraphics[width=.15\linewidth]{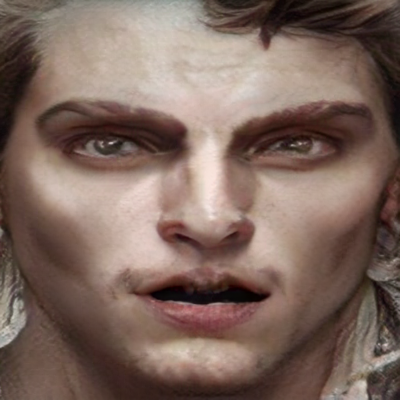} & \hspace{0.1cm}
     \includegraphics[width=.15\linewidth]{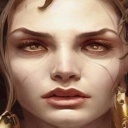} &
     \includegraphics[width=.15\linewidth]{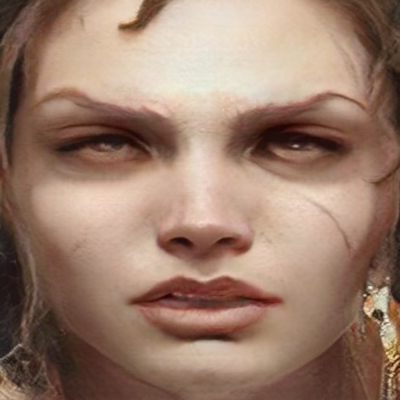} & 
     \includegraphics[width=.15\linewidth]{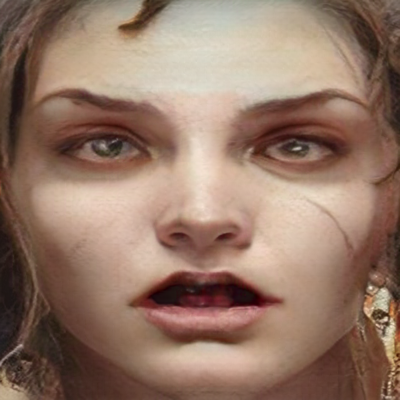}  \\

     \includegraphics[width=.15\linewidth]{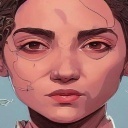}  & 
     \includegraphics[width=.15\linewidth]{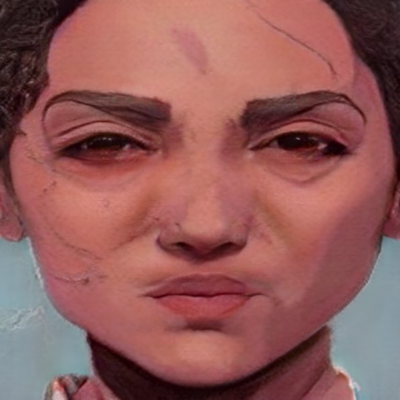} &
     \includegraphics[width=.15\linewidth]{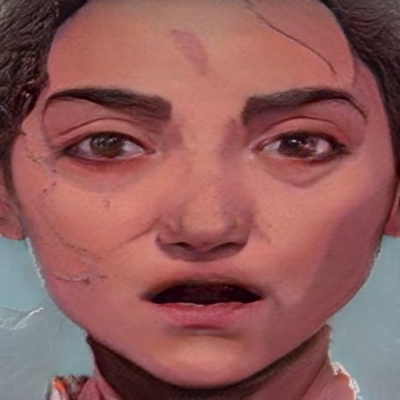} & \hspace{0.1cm}
     \includegraphics[width=.15\linewidth]{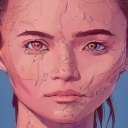} &
     \includegraphics[width=.15\linewidth]{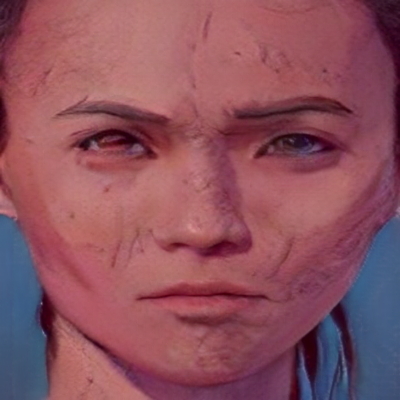} & 
     \includegraphics[width=.15\linewidth]{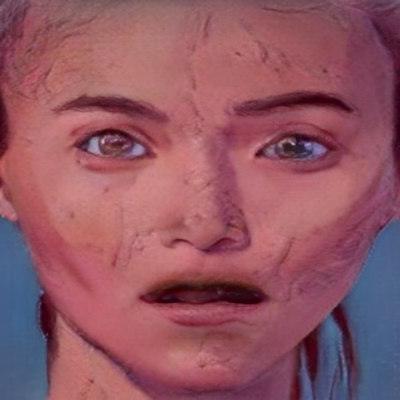}  \\

     \includegraphics[width=.15\linewidth]{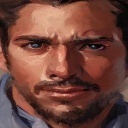}  & 
     \includegraphics[width=.15\linewidth]{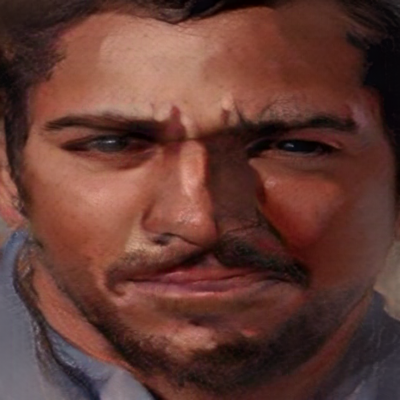} &
     \includegraphics[width=.15\linewidth]{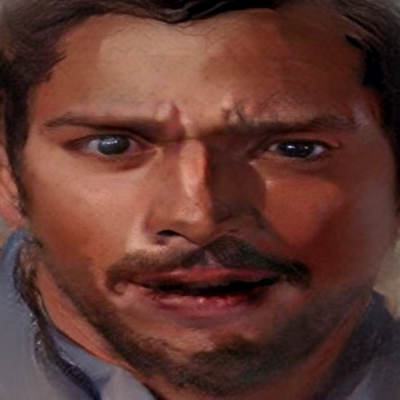} & \hspace{0.1cm}
     \includegraphics[width=.15\linewidth]{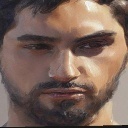} &
     \includegraphics[width=.15\linewidth]{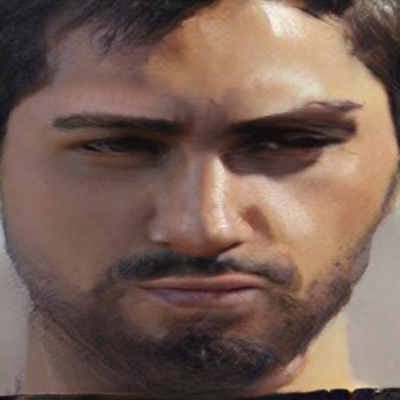} & 
     \includegraphics[width=.15\linewidth]{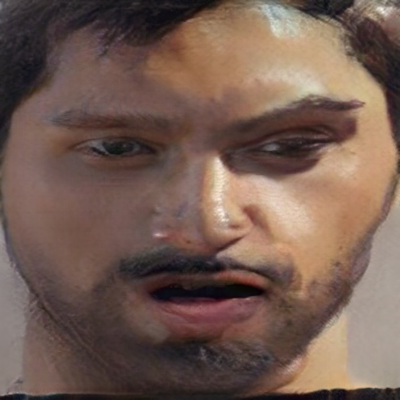}  \\

     \includegraphics[width=.15\linewidth]{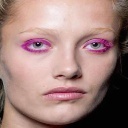}  & 
     \includegraphics[width=.15\linewidth]{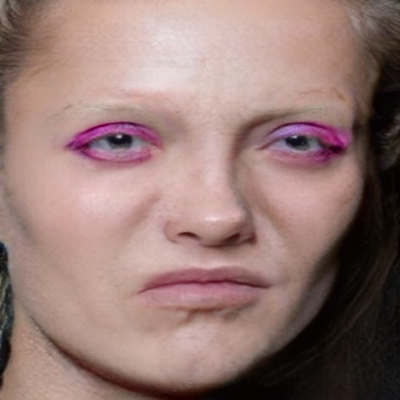} &
     \includegraphics[width=.15\linewidth]{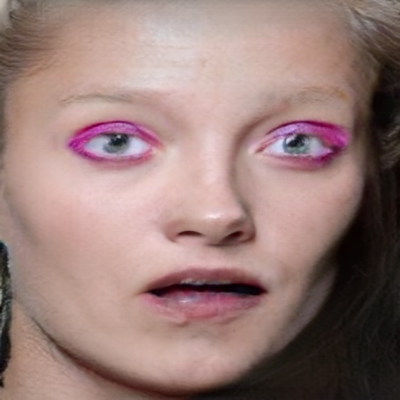} & \hspace{0.1cm}
     \includegraphics[width=.15\linewidth]{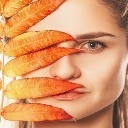} &
     \includegraphics[width=.15\linewidth]{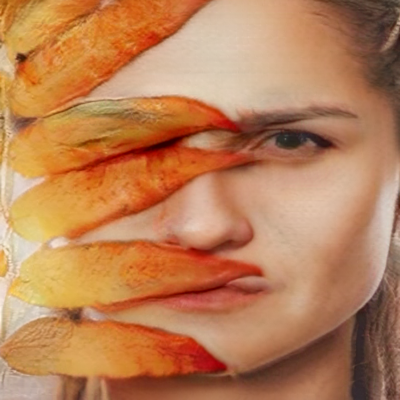} & 
     \includegraphics[width=.15\linewidth]{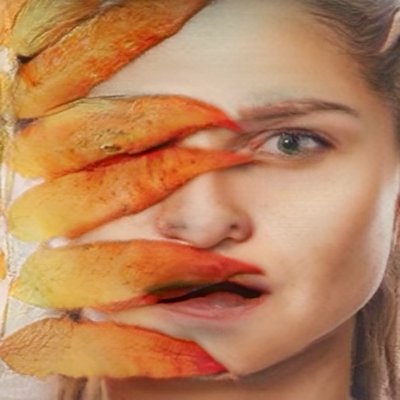}  \\
    \end{tabular}
    \caption{Results of the proposed method on diverse facial images: Row 1: faces with non-uniform, multi-colored illumination. Row 2: faces in the style of fantasy artist Peter Mohrbacher. Row 3: faces in the style of artist Tomer Hanuka. Row 4: impasto in the style of Gregory Manchess. Row 5: faces with makeup and occlusion. All input images were generated using Stable Diffusion \cite{rombach2022high}. Results demonstrate strong generalization of the proposed method, that synthesizes realistic facial expressions while preserving the artistic details and styles of the input images.}
    \label{fig:mix-lighting-results}
\end{figure}

\begin{figure}[t]
  \centering
  \fontsize{9}{9}\selectfont
  \renewcommand{\arraystretch}{1}
  \setlength{\tabcolsep}{0pt}
  \begin{tabular}{cccccc}
     Input & Happy & Input & Happy & Input & Happy\\    
     \includegraphics[width=.15\linewidth]{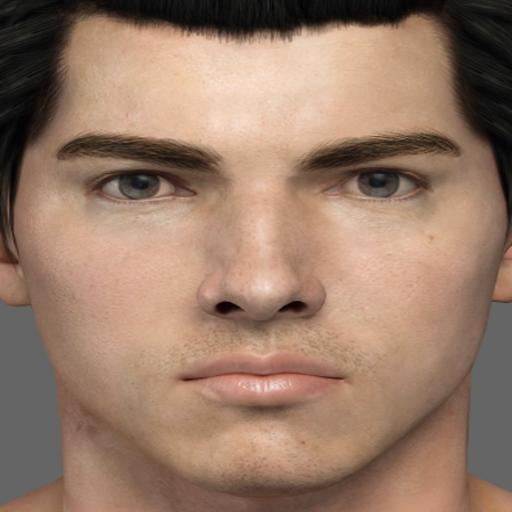}  & 
     \includegraphics[width=.15\linewidth]{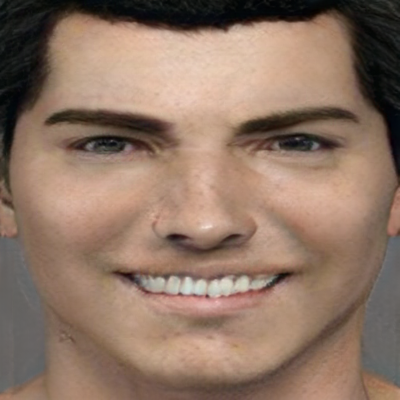} & \hspace{0.1cm}
     \includegraphics[width=.15\linewidth]{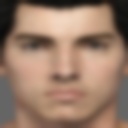} &
     \includegraphics[width=.15\linewidth]{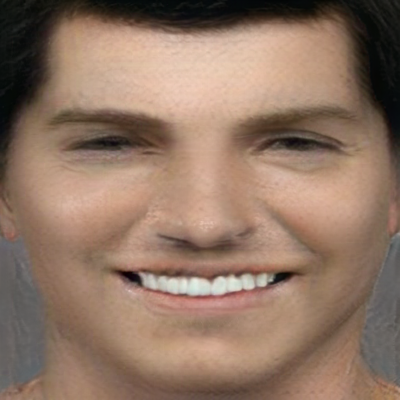} & \hspace{0.1cm}
     \includegraphics[width=.15\linewidth]{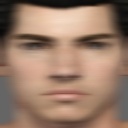} & 
     \includegraphics[width=.15\linewidth]{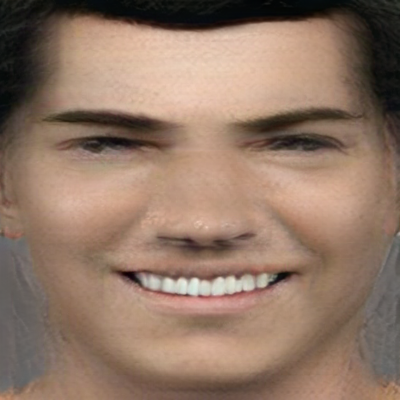} \\

    \includegraphics[width=.15\linewidth]{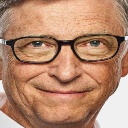}  & 
     \includegraphics[width=.15\linewidth]{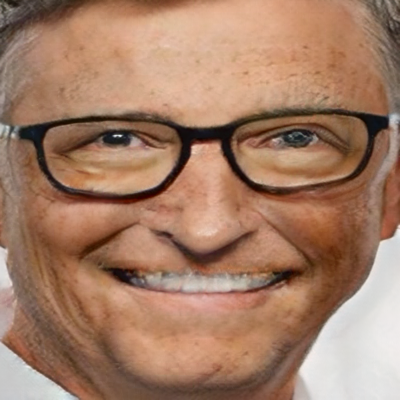} & \hspace{0.1cm}
     \includegraphics[width=.15\linewidth]{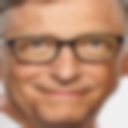} &
     \includegraphics[width=.15\linewidth]{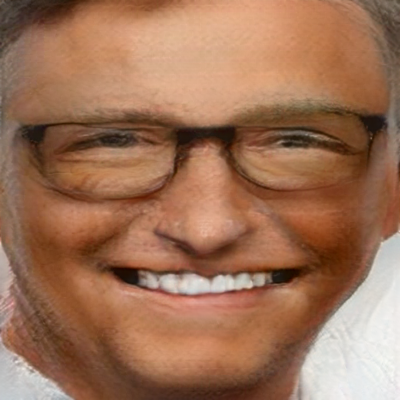} & \hspace{0.1cm}
     \includegraphics[width=.15\linewidth]{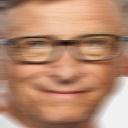} & 
     \includegraphics[width=.15\linewidth]{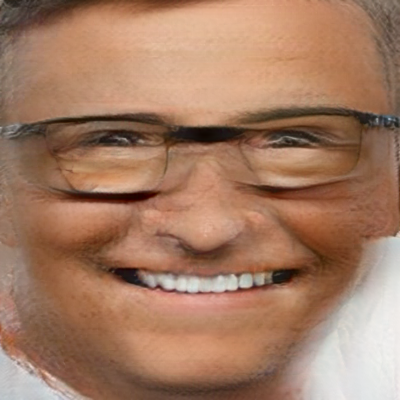} \\

    \includegraphics[width=.15\linewidth]{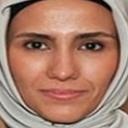}  & 
     \includegraphics[width=.15\linewidth]{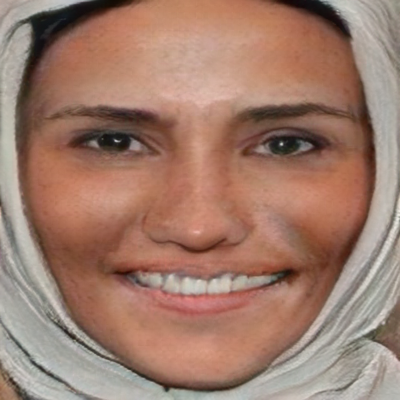} & \hspace{0.1cm}
     \includegraphics[width=.15\linewidth]{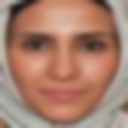} &
     \includegraphics[width=.15\linewidth]{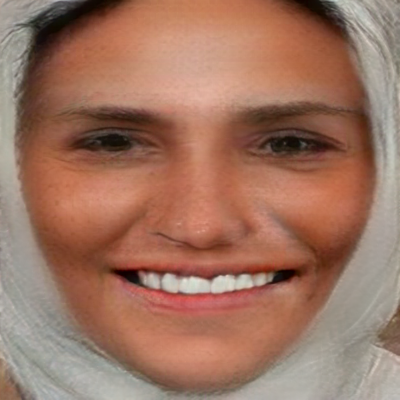} & \hspace{0.1cm}
     \includegraphics[width=.15\linewidth]{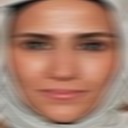} & 
     \includegraphics[width=.15\linewidth]{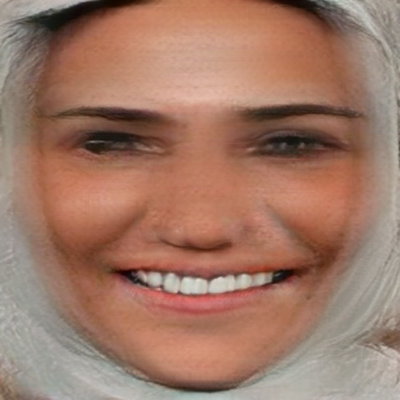} \\
     \multicolumn{2}{c}{Original} &
     \multicolumn{2}{c}{Gaussian-smoothed} &
     \multicolumn{2}{c}{Motion-blurred} \\
\end{tabular}
\caption{RegGAN results on Gaussian-smoothed and motion-blurred images. While the expression layer successfully induces desired expressions on sharp as well as blurred inputs, the sharpening effect of our refinement network causes inputs and outputs to have different levels of blurriness. This can be beneficial in some contexts, but not so desirable in others.
}
\label{fig:blur-comparison-fig}
\end{figure}

\emph{Improved Expression Generalization via Regression:}
Figures \ref{fig:comp_with_state_of_the_art_happy_exp} and \ref{fig:reggan_comparison4} show that GAN-based methods often struggle to generalize across diverse image domains, as they attempt to model the full data distribution, including both identity and expression variations. In contrast, the regression-based approach, MR, focuses on learning expression biases from the data, enabling it to consistently introduce realistic expressions across different images. By combining these two in RegGAN, regression ensures accurate expression synthesis, while the GAN component enforces realism. This hybrid design generalizes across human portraits, avatars, cartoons, as well as statues (see Figure~\ref{fig:out-of-dist-celeb}). 
As demonstrated in Figure~\ref{fig:mix-lighting-results}, RegGAN generalizes well over a diverse range of facial images with varying illuminations, artistic styles, and occlusions. Figure~\ref{fig:blur-comparison-fig} shows results on Gaussian-smoothed and motion-blurred images. While the expression layer successfully induces desired expressions on sharp as well as blurred inputs, the sharpening effect of our refinement network reduces the original blurriness of the input. This can be beneficial in some contexts and not desirable in others. The sharpness is due to an implicit prior that input and output images are sharp. When they are not sharp, the refinement network forcibly introduces sharpness. Future work should address blur-preserving FES. 

\emph{Dependency on Paired Expressions:}
Although RegGAN generalizes well on a variety of images, it relies on paired data consisting of neutral and target expressions to train the regression layer. This dependency may restrict its use when such data is unavailable. Future work could explore unpaired or weakly supervised alternatives.

\begin{figure}[h]
    \centering  
    \includegraphics[width=\linewidth]{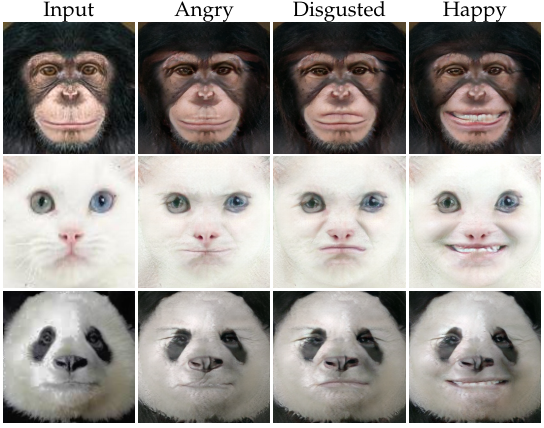}
    \caption{RegGAN produces human-like features, including eyes and facial contours, in cat and panda images. While cats naturally have vertical pupils, RegGAN tends to make them circular due to its training on human faces.}   
    \label{fig:failure_cases}
\end{figure}

\emph{Computational Cost of the Refinement Network:} The refinement network in RegGAN enhances photorealism but introduces significant computational overhead. Results in Table~\ref{tab:quant-analysis-ablation} show that the latent attention blocks and the hourglass network \cite{newell2016stacked} used in the attention unit improve performance, but they also add a large number of parameters, making the model unsuitable for real-time inference or deployment on resource-constrained devices. This highlights a fundamental trade-off: more depth yields better realism but reduces efficiency. In the future, the hourglass network in the attention unit could be replaced with lightweight alternatives to reduce computational cost without sacrificing realism.

\section{Failure Cases}
\label{sec:failure_cases}
In Figure~\ref{fig:failure_cases}, we present the results of RegGAN on animal faces. Row 1 shows that our proposed method synthesizes a realistic happy expression on a monkey's face. However, it fails to produce convincing angry and disgusted expressions on the same face. Rows 2 and 3 show that RegGAN generates realistic human-like angry, disgusted, and happy expressions on cat and panda images. Since the model is trained on human faces, it produces human-like eyes and facial contours in the synthesized cat and panda images. Although it induces convincing expressions and preserves facial details, the anthropomorphic bias limits the overall quality of the results.

\section{Conclusion}
\label{sec:conclusion}
In this work, we evaluate the generalization ability of existing GANs, including StarGAN, GANimation, DAI2I, SARGAN, US-GAN, and MR, on out-of-distribution images. These GANs are trained on real human facial expression datasets and tested on human faces, portraits, avatars, statues, and animal faces outside the training distribution. Our observations indicate that their generalization to such images is limited. This limitation arises because GANs attempt to learn the high-dimensional distributions of input and output images, as well as mapping between them, while the training datasets sparsely cover these distributions. Therefore, what these GANs learn is only an estimate of the actual distribution. As a result, when these GANs are presented with input images that are distant from the learned distributions, they fail to map these images to the correct region of the target distribution. In this work, we propose a solution to this problem: first, the input images are translated to an intermediate representation, and then this intermediate representation is mapped to the final output distribution. The first translation is performed by a ridge regression layer learned through localized receptive fields, while the second translation is achieved by optimizing the adversarial loss. The proposed RegGAN demonstrates significantly better generalization capability compared to other GANs.

\bibliographystyle{unsrt}  
\bibliography{references}

\end{document}